\documentclass[11pt,a4paper]{article}
\pdfoutput=1
\usepackage[hyperref]{emnlp-ijcnlp-2019}
\usepackage{times}
\usepackage{latexsym}

\usepackage{amsmath}
\usepackage{amsfonts}
\usepackage{CJKutf8}
\newcommand*{\Ja}[1]{%
\begin{CJK}{UTF8}{ipxm}#1\end{CJK}}
\usepackage{linguextwocolumn}
\usepackage{cgloss4e}
\newcommand{\agls}[1]{\textsc{\scriptsize #1}}
\usepackage{tikz}
\newcommand\myline[1][]{%
   \,\tikz[baseline]\draw[very thick,#1](0,-\dp\strutbox)--(0,\ht\strutbox);\,%
}
\usepackage{tcolorbox}

\aclfinalcopy % Uncomment this line for the final submission
\title{On the Definition of Japanese Word}

\author{Yugo Murawaki\\
Graduate School of Informatics, Kyoto University\\
Yoshida-honmachi, Sakyo-ku, Kyoto, 606-8501, Japan \\
\texttt{murawaki@i.kyoto-u.ac.jp} }

\date{}

\begin{document}
\maketitle

\begin{abstract}
The annotation guidelines for Universal Dependencies (UD) stipulate that the basic units of dependency annotation are \emph{syntactic words}, but it is not clear what are syntactic words in Japanese.
Departing from the long tradition of using phrasal units called \emph{bunsetsu} for dependency parsing, the current UD Japanese treebanks adopt the Short Unit Words.
However, we argue that they are not syntactic word as specified by the annotation guidelines.
Although we find non-mainstream attempts to linguistically define Japanese words, such definitions have never been applied to corpus annotation.
We discuss the costs and benefits of adopting the rather unfamiliar criteria.
\end{abstract}

\section{Introduction}

Japanese occupies a unique position in text processing because unlike Chinese and Thai, it is morphologically rich and yet does not use white space to delimit words.
It is no wonder that Japanese language processing (JLP) has developed its own peculiar task designs.
To facilitate text processing in multilingual settings~\citep{CoNLL2006shared}, however, Japanese must be analyzed in a manner comparable to other languages.

The Universal Dependencies (UD) project~\citep{McDonald:ACL2013} provides a rare opportunity to overhaul grammatical treatments of Japanese because otherwise the research area is considered too mature to change.
Here we focus on the lowest layer of corpus annotation, that is, the definition of word.
Departing from the long tradition of using phrasal units called \emph{bunsetsu} for dependency parsing~\citep{Kurohashi:SNL1994knp},
Universal Dependencies for Japanese (UD Japanese)~\citep{Tanaka:LREC2016,Asahara:LREC2018} adopts short, morpheme-like units called the Short Unit Words (SUWs)~\citep{Maekawa:LRE2014}.

Since UD considers morphology to be word-internal domains, UD Japanese clearly violates the current UD guidelines.
However, we argue that SUWs are an understandable tentative solution.
To define UD-compliant words, we need to classify dependent (function) morphemes into (1) affixes that are part of other words and (2) clitics that are treated as separate words~\citep{UD-word-def}.
Although there are attempts to apply the distinction to Japanese~\citep{Hattori:1960fuzoku,Miyaoka:2015go}, they remain virtually unknown to JLP and Japanese corpus studies.

After reviewing word-like units that have been used in JLP and and Japanese corpus studies,
we show that, as exemplified in Figure~\ref{fig:proposal}, if we approximate Japanese morphemes by SUWs, it is possible to create syntactic words by combining one or more SUWs.
However, we expect the cost of newly creating workable annotation guidelines to outweigh the benefit of enabling cross-linguistic comparison.

Note that \citet{Pringle:2016word} provides an excellent review on the definition of word and UD Japanese.
We urge readers to read it first.
Here we try to complement it but we may still miss some important information since we are connecting segregated research fields.
We hope that this article will serve as a basis for future discussion.

\begin{figure*}[tb]
 \centering
\begin{tcolorbox}[colback=white]
\setlength{\Exindent}{0.0em}%
\setlength{\Exlabelwidth}{0.0em}%
\setlength{\Exlabelsep}{0.0em}%
\exg.[]
  \Ja{彼女} \Ja{の} \Ja{美し} \Ja{さ} \Ja{が} \Ja{失わ} \Ja{れ} \Ja{ない} \\
  she =\agls{GEN} beautiful -\agls{NLZ} =\agls{NOM} lose -\agls{PASS} -\agls{NEG.NPST} \\
  `Her beauty is not lost'

\begin{tabular}{@{\hspace{0.0em}} l ccccccccccccccccc @{\hspace{0.0em}}}
SUWs:
 & \textbar\, & \Ja{彼女} & \textbar\, & \Ja{の} & \textbar\, & \Ja{美し} & \textbar\, & \Ja{さ} & \textbar\, & \Ja{が} & \textbar\, & \Ja{失わ} & \textbar\, & \Ja{れ} & \textbar\, & \Ja{ない} & \textbar \\
\textbf{Syn. words}:
 & \textbar\, & \Ja{彼女} & \textbar\, & \Ja{の} & \textbar\, & \Ja{美し} && \Ja{さ} & \textbar\, & \Ja{が} & \textbar\, & \Ja{失わ} && \Ja{れ} && \Ja{ない} & \textbar \\
LUWs:
 & \textbar\, & \Ja{彼女} & \textbar\, & \Ja{の} & \textbar\, & \Ja{美し} & \textbar\, & \Ja{さ} & \textbar\, & \Ja{が} & \textbar\, & \Ja{失わ} & \textbar\, & \Ja{れ} & \textbar\, & \Ja{ない} & \textbar \\
\emph{Bunsetsu}:
 & \textbar\, & \Ja{彼女} && \Ja{の} & \textbar\, & \Ja{美し} && \Ja{さ} && \Ja{が} & \textbar\, & \Ja{失わ} && \Ja{れ} && \Ja{ない} & \textbar \\
\end{tabular}

\end{tcolorbox}

 \caption{Japanese syntactic words as concatenations of SUWs (Short Unit Words).
A glossed example is followed by various segmentations.
The vertical bars indicate unit boundaries.
We can see that syntactic words (2nd row) are different from existing units such as LUWs (Long Unit Words) (3rd row) and \emph{bunsetsu} (last row).
}
 \label{fig:proposal}
\end{figure*}

\section{UD Annotation Guidelines on Words} \label{sec:ud-policy}

UD treats words as the atoms of syntactic analysis and calls them \emph{syntactic words}.
The guidelines on tokenization and word segmentation devote much space to distinguishing syntactic words from orthographic and phonological words~\citep{UD-word-def}.
However, there are no orthographic words in Japanese and phonological words are irrelevant to JLP.

In our understanding, the key assumption behind syntactic words is that there is a boundary between morphology and syntax.
Although the guidelines do not make it explicit, this entails the manual task of classifying dependent morphemes into affixes and clitics.
The former are part of words while the latter are words by themselves.

UD's decision is based on the lexicalist hypothesis that grew out of the early stages of generative grammar~\citep{Chomsky:1970}.
It is important to note that this hypothesis does not necessarily have strong theoretical support.
There are two extreme positions: one recognizes morphology and syntax as distinct domains~\citep{DiSciullo:1987}
and the other denies the autonomy of morphology and incorporates morphology into syntax~\citep{Lieber:1992}.
UD appears to lean toward the former but most theoretical analyses fall somewhere in between.
Moreover, even if the autonomy of morphology is accepted to a certain degree, context-free grammar (CFG) trees are often employed to analyze word formations~\citep{Kageyama:1993EN}.

\section{\emph{Bunsetsu} in Traditional JLP} \label{sec:jlp}

The traditional JLP pipeline is morphological analysis, \emph{bunsetsu} chunking, and dependency parsing in this order.
What is conventionally referred to as morphological analysis in JLP is the joint task of segmentation, part-of-speech (POS) tagging and lemmatization.
Morphemes, often simply referred to as ``words'' in the literature, are not necessarily morphemes in the sense of linguistics.
In the standard grammar-and-dictionary approach, subdividing text into smaller productive elements improves the dictionary's coverage while adding longer fixed elements (e.g., proper nouns and unproductive compounds) to the dictionary mitigates errors during analysis.
Morphemes are the outcome of engineering decisions to achieve a balance between the two.
For this reason, morphemes are not explicitly defined in, for example, the Kyoto University Text Corpus~\citep{Kurohashi1998full} but the default dictionary of the morphological analyzer JUMAN~\citep{Kurohashi:SNL1994juman} serves as the de facto criteria.
The resulting lack of consistency in segmentation is criticized by corpus linguists~\citep{Maekawa:LRE2014}.

\emph{Bunsetsu} chunking is usually performed as a preprocessing of dependency parsing.
Originating from the Japanese grammatical tradition~\citep{Hashimoto:1933EN}, \emph{bunsetsu} is a relatively stable\footnote{
Most disagreements come from treatment of grammaticalized elements.
} phrasal unit, typically consisting of one or more \emph{content morphemes} followed by zero or more \emph{function morphemes}.\footnote{To be precise, there are a small number of prefixes and proclitics in Japanese.
In \emph{bunsetsu} chunking, they are attached to content morphemes.
}
Note here that there is no distinction between affixes and clitics.
In example~\Next, both the nominalizer suffix \Ja{さ} and the case enclitic \Ja{が} are part of the same \emph{bunsetsu} ($||$ denotes \emph{bunsetsu} boundaries).
\exg.
  $||$ \Ja{美し} \Ja{さ} \Ja{が} $||$ \Ja{失わ} \Ja{れ} \Ja{ない} $||$ \\
  {} beautiful -\agls{NLZ} =\agls{NOM} {} lose -\agls{PASS} -\agls{NEG.NPST} {}
\\ \label{ex:beauty}

After \emph{bunsetsu} chunking, dependency relations are assigned to \emph{bunsetsu} pairs.\footnote{
The CoNLL-2009 shared task on syntactic and semantic dependencies~\citep{Hajic:CoNLL2009shared}
adopted morpheme-based dependency trees for Japanese.
These trees were constructed by automatically dissolving \emph{bunsetsu} chunks.
}

\section{NINJAL's Units} \label{sec:ninjal}

The SUWs, which were chosen as the atoms of syntactic relations by UD Japanese, have come out of more than six decades of corpus studies by the National Institute for Japanese Language and Linguistics (NINJAL).
It is worth noting that although gradually adopting NLP technologies, corpus linguistics remains culturally very different from NLP.
For example, tedious manual labor is often preferred over less reliable but orders of magnitude faster automatic analysis.
Given the fact that parsing has become so mature that the focus of corpus building in JLP has shifted to zero anaphora, discourse and other higher-level structures~\citep{Hangyo:PACLIC2012,Kawahara:COLING2014}, NINJAL is perhaps the only institution in Japan that can reinvent the wheel.

SUWs are one of several units created by NINJAL.
Other units relevant to the present discussion are LUWs (Long Unit Words) and \emph{bunsetsu}.\footnote{
The remaining units are Middle Unit Words (designed for speech research) and Minimal Unit Words (used to create SUWs).}
The predecessors of SUWs and LUWs were created out of an urgent need to count word-like units in text in a consistent way.
When defining these units, NINJAL declared itself agnostic with theoretical linguistics and maintained what it called an operationalist approach.
This resulted in book-length guldelines with literally hundreds of rules~\citep{BCCWJ-word:2006EN} but none of them were justified on linguistic grounds.\footnote{
We previously worked on acquiring unknown morphemes (those missing from the dictionary) from text~\citep{Murawaki:EMNLP2008full}.
While our system operated in conjunction with the default dictionary of the morphological analyzer JUMAN~\citep{Kurohashi:SNL1994juman}, we explored the possibilities of making it BCCWJ-compatible.
However, we gave up the plan because the guidelines were too complex for a computer to comply with.
}
They sometimes lead to counter-intuitive outcomes.
For example, \Ja{寿司屋} (sushi shop) is one SUW but \Ja{ラーメン\myline[dotted]屋} (noodle shop) is two SUWs (\myline[dotted] marks SUW boundaries in in-line examples).
What NINJAL emphasizes is that as long as we adhere to the rules, the number of SUWs is counted consistently.

SUWs, LUWs and \emph{bunsetsu} have inclusion relations in this order.
A \emph{bunsetsu} contains one or more LUWs, and an LUW in turn contains one or more SUWs.
SUWs are morpheme-like\footnote{
Although SUWs are generally shorter than or equal to what are treated as words in traditional JLP,
an SUW is created by concatenating one or more Minimal Unit Words.
Thus Minimal Unit Words are closer to morphemes.
} while LUWs are sometimes longer.
The main differences between SUWs and LUWs lie in treatment of compound nouns and functional expressions.
In example~\Next, \Ja{魚\myline[dotted]フライ} (fish fry) forms one LUW (\myline indicates LUW boundaries).

\exg.
  $||$ {\myline} \Ja{魚} \Ja{フライ} {\myline} \Ja{だけ} {\myline} \Ja{を} {\myline} $||$ \\
  {} {} fish fry {} =only {} =\agls{ACC} {} {}
\\ \label{ex:luw-noun}

Dependent SUWs are mostly kept apart.

\exg.
  $||$ {\myline} \Ja{美し} {\myline} \Ja{さ} {\myline} \Ja{が} {\myline} $||$ \\
  {} {} beautiful {} -\agls{NLZ} {} =\agls{NOM} {} {}
\\ \label{ex:luw-deriv-adj}

\exg.
  $||$ {\myline} \Ja{失わ} {\myline} \Ja{れ} {\myline} \Ja{ない} {\myline} $||$ \\
  {} {} lose {} -\agls{PASS} {} -\agls{NEG.NPST} {} {}
\\ \label{ex:luw-deriv-verb}

You can see that the elaborate system has no room for the distinction between affixes and clitics.

NINJAL's flagship corpus, the Balanced Corpus of Contemporary Written Japanese (BCCWJ)~\citep{Maekawa:LRE2014}, is annotated with the SUWs, LUWs and \emph{bunsetsu}.
Other corpora from NINJAL such as a corpus of Meiji-Taisho-era Japanese~\citep{Ogiso:DH2017} generally follow the pattern.
Note that NINJAL itself chose \emph{bunsetsu}-based dependencies for annotating the BCCWJ~\citep{Asahara;ALR2016}.

\section{Short Unit Words in UD Japanese} \label{sec:ud-japanese}

The UD Japanese team decided to create Japanese resources through automatic rule-based conversion from the BCCWJ and BCCWJ-compatible treebanks.\footnote{
Adaptation to the BCCWJ schema is to be performed first if existing treebanks are not BCCWJ-compatible.
}
Thus, UD Japanese corpora have no way of providing information not present in the original treebanks unless rules are rich enough to inject external knowledge.

UD Japanese adopts SUWs as the units of syntactic annotation.
One of the initially planned use cases for UD Japanese was tree-based statistical machine translation (SMT)~\citep{Hoshino:IJCNLP2013} although it quickly got obsolete with the rise of end-to-end neural machine translation~\citep{Sutskever:NIPS2014,Bahdanau:ICLR2014}, which outperforms SMT without using any explicit syntactic information.
To facilitate Japanese-to-English translation, Japanese syntactic trees were desired to be represented such that they could be mapped to those of the isolating language more transparently.
Obviously, this is not a linguistically-sound motivation but is understandable given that the last of UD's six goals is to ``support well downstream language understanding tasks (relation extraction, reading comprehension, machine translation, \textellipsis)''~\citep{UD-introduction}.

The adoption of SUWs clearly violates the current UD guidelines that are based on the lexicalist hypothesis~\citep{UD-word-def}.
We could resort to a weaker version of the hypothesis that differentiates inflectional and derivational morphology and recognizes syntactic nature of inflection.
Still, we have no choice but to treat derivational affixes as part of words.
\Ja{美し\myline[dotted]さ} (beauty) in example~\ref{ex:beauty} is treated as two words, given POS tags, \texttt{ADJ} and \texttt{PART} (particles), and connected with the dependency relation \texttt{mark} (finite clause subordination).
However, the suffix \Ja{さ} derives the noun from the adjective and thus should be part of the single word with POS tag \texttt{NOUN}.

\section{Distinguishing Clitics from Affixes} \label{sec:miyaoka}

As we have seen, JLP and Japanese corpus studies have not made a distinction between affixes and clitics.
It is because in traditional Japanese \emph{school grammar}, word-like elements are first classified according to the content/function distinction.
The latter are then divided into conjugable \emph{jod\={o}shi} and non-conjugable \emph{joshi}~\citep{Hashimoto:1933EN}, which are orthogonal to the affix/clitic distinction.\footnote{
\citet{Miyaoka:2015go} attributes the absence of the affix/clitic distinction in Japanese grammatical tradition to the peculiarity of the Japanese writing system, in which logographic \emph{kanji} are used to write content words while syllabic \emph{hiragana} follow \emph{kanji} to represent inflectional endings and other functional elements.
He suggests that the content/function dichotomy obscures the formal distinction between affixes and clitics.
}
\citet{Pringle:2016word} gives a thorough review of grammatical studies of Japanese in this regard.

However, this does not mean that the distinction has never been applied to Japanese.
In fact, the structuralist linguist Shiro Hattori presented a formal classification in which the primary division is between free forms (\emph{jiy\={u}-keishiki}) and bound forms (\emph{fuzoku-keishiki})~\citep{Hattori:1960fuzoku}.
Minimal free forms, or words, are divided into independent words (\emph{jiritsu-go}) and bound words (\emph{fuzoku-go}).
Bound words and bound forms correspond to clitics and affixes, respectively, in our terminology.
He then presented three principles for identifying bound words (clitics), and as by-products, bound forms (affixes).
Unfortunately, while he demonstrated the general applicability of the principles using examples not only from Japanese but from English, Russian, Turkish and other languages, his discussion on Japanese was so limited that it can hardly serve as a starting point of corpus annotation.

Recently, Osahito Miyaoka, a field linguist working on the polysynthetic language Central Alaskan Yup'ik, published a monograph on the definition of word, with a strong structuralist flavor~\citep{Miyaoka:2015go}.
Unlike \citet{Hattori:1960fuzoku}, he presented his own classification that covered major affixes and clitics (Table~2 of \citet{Miyaoka:2015go}) and discussed borderline cases in detail.
Although dependency grammar is clearly out of the scope of his analysis, we think that \citet{Miyaoka:2015go} can be a feasible basis for corpus annotation.

\section{Feasibility Assessment} \label{sec:feasibility}

In this section, we examine the possibility of adopting Miyaoka's classification for UD Japanese.
This basically implies additional rule writing because the BCCWJ and BCCWJ-compatible treebanks are too big to discard.
Needless to say, a substantial amount of manual work is needed to extend his classification to missing elements.

Our preliminary examination suggests that the affix/clitic distinction can be made at the level of lexical items.
For example, the negation marker \Ja{ない} is a suffix when following a verb and an enclitic when negating an adjective.

\exg.
  $||$ \Ja{書か} \Ja{ない} $||$ \\
  {} write -\agls{NEG}.\agls{NPST} {}
\\ \label{ex:nai-verb}

\exg.
  $||$ \Ja{寒く} \Ja{ない} $||$ \\
  {} cold.\agls{ADV} =\agls{NEG}.\agls{NPST} {}
\\ \label{ex:nai-adj}

There are one syntactic word in example~\ref{ex:nai-verb} and two words in example~\ref{ex:nai-adj}.
However, these two can easily be distinguished because they are given different POS tags.

One known exception is \Ja{らしい}.
Depending on the context, it can be a derivational suffix (-ly) or an enclitic (seem to be).

\exg.
  $||$ \Ja{とても} $||$ \Ja{男} \Ja{らしい} $||$ \Ja{人} $||$ \\
  {} very {} man -\agls{ADJLZ}.\agls{NPST} {} person {}
\\ \label{ex:rashii-suffix}

\exg.
  $||$ \Ja{犯人} \Ja{は} $||$ \Ja{どうやら} $||$ \Ja{男} \Ja{らしい} $||$ \\
  {} culprit =\agls{TOP} {} apparently {} man =seem.\agls{NPST} {}
\\ \label{ex:rashii-enclitic}
\vspace{-2.0em} %% kill mysterious large space

However, no disambiguation is made in the BCCWJ because it follows nominals in both cases.
We have no choice but to disambiguate every token in text.

From a processing point of view, our proposal entails a novel task in JLP: segmentation into syntactic words.
Given the existing ecosystem, the straightforward way to do this is the pipeline of (1) segmentation into SUWs and (2) chunking of SUWs into syntactic words.
However, the very motivation behind UD is to build a single system that works for any language, and the community would not favor such a language-specific solution.
The direct identification of syntactic words is another option, but we anticipate a loss in performance.\footnote{
JLP experiences a \emph{great reset} with the introduction of UD.
Given that high-coverage dictionaries have traditionally served as powerful resources to achieve over 99\% F-measure in segmentation~\citep{Kudo:EMNLP2004full,Maekawa:LRE2014}, drastically lower performance of dictionary-less universal systems~\citep{Straka:CoNLL2017} is remarkable although the scores are not directly comparable.

Dictionaries remain key ingredients in JLP.
\citet{Tolmachev:NAACL2019} managed to create a dictionary-less (non-UD) morphological analyzer that performed comparably with a dictionary-based one.
However, they distilled dictionary knowledge through a huge raw corpus automatically analyzed by the dictionary-based anazyler.
}

The elimination of affixes from syntactic analysis should be compensated in some way or other.
Some of their properties can be reflected by modifying the POS tags, but the rest are to be described as (linguistic) features.
This is not an easy task not only because UD Japanese currently does not use linguistic features at all but also because they are new to the Japanese dependency parsing community as a whole.
Thanks to a high degree of form-meaning transparency of the agglutinative language, it makes direct use of lexical items as machine learning features~\citep{Uchimoto:EACL1999} and does not bother to map them to linguistic features.

\section{Discussion and Conclusions} \label{sec:discussion}

As described in Section~\ref{sec:ud-japanese}, UD Japanese resources are created through automatic conversion from existing treebanks.
This is partly because the team considers that UD guidelines are not yet stabilized but subject to future changes that may be responses to new, typologically unfamiliar language or may simply come from improved awareness of typological studies~\citep{Croft:TLT15}, where inherent difficulties in cross-linguistic comparison are widely recognized~\citep{Haspelmath:Language2010}.
While automatic conversion allows us to rapidly adapt to changes in the guidelines, all necessary information must be present in the original treebanks.

In order to comply with the current UD guidelines on syntactic words, we have to add another layer of annotation to existing treebanks because they lack the distinction between affixes and clitics.
The anticipated non-negligible costs lead us to question whether the manual work pays.

One area that is likely to benefit is cross-linguistic comparison (UD's goal 2).
Suppose that we apply cross-lingual projection of parsing models~\citep{Cohen:EMNLP2011} and find that a certain pair of languages are (dis)similar.
Naturally, we want to ensure that it reflects structural properties of languages, not arbitrary design decisions on corpus annotation.
However, the goal can be achieved only when the same level of care is given to other languages.
Besides, the distinction between affixes and clitics itself is known to be fragile.
In fact, after a detailed examination, \citet{Haspelmath:FLin2011,Haspelmath:CatCat2015} concluded that there were no good criteria for defining syntactic words as a cross-linguistically valid concept.

Another area to consider is downstream tasks (UD's goal 6).
Unless text generation is involved, it hardly matters whether a certain function is realized by morphological or syntactic devises.
Given the semantic transparency in the agglutinative language, unified treatment of affixes and clitics appears to have an advantage in formal semantics~\citep{Bekki:2010EN}, for example.

Rather, applications cast doubt on UD's current framework.
Although the affix/clitic distinction is not clear-cut, subtle variations in design result in very different representations.
This also reminds us of \citet{Croft:TLT15}'s proposal for construction-based annotation.
For example, English uses a copula for \emph{predicate nominal constructions} while Russian does not, and by directly linking content words, UD provides comparable representations.
Why not applying the same logic to the morphology-syntax division?

To conclude, we found that with manual work, UD Japanese would likely be able to comply with the current UD guidelines on words, but we are unsure if it is the right course of action to take.

\bibliography{paper}
\bibliographystyle{acl_natbib}

\end{document}